\documentclass[letterpaper, 10pt, conference]{ieeeconf}  % Comment this line out if you need a4paper

\IEEEoverridecommandlockouts                              % This command is only needed if 
\usepackage{amsmath,mathtools}
\usepackage{array}
\usepackage{amssymb}
\usepackage{caption}
\usepackage{subcaption}
\usepackage{graphics} % for pdf, bitmapped graphics files
\usepackage{epsfig} % for postscript graphics files
\DeclareMathOperator*{\argmaxC}{\arg\max}   % rbp

\makeatletter
\let\NAT@parse\undefined
\makeatother
\usepackage{hyperref}

\title{\LARGE \bf
Learning Human-like Hand Reaching for Human-Robot Handshaking}

\author{Vignesh Prasad$^{1}$, Ruth Stock-Homburg$^{1}$, Jan Peters$^{1,2}$% <-this % stops a space
\\ {\tt\small \{vignesh.prasad,ruth.stock-homburg,jan.peters\}@tu-darmstadt.de}
\thanks{$^{1}$Technical University of Darmstadt, Germany}%
\thanks{$^{2}$Max Planck Institute for Intelligent Systems, T\"ubingen, Germany}%
}

\begin{document}

\maketitle
\thispagestyle{empty}
\pagestyle{empty}

%%%%%%%%%%%%%%%%%%%%%%%%%%%%%%%%%%%%%%%%%%%%%%%%%%%%%%%%%%%%%%%%%%%%%%%%%%%%%%%%
\begin{abstract}

One of the first and foremost non-verbal interactions that humans perform is a handshake. It has an impact on first impressions as touch can convey complex emotions. This makes handshaking an important skill for the repertoire of a social robot. In this paper, we present a novel framework for learning reaching behaviours for human-robot handshaking behaviours for humanoid robots solely using third-person human-human interaction data. This is especially useful for non-backdrivable robots that cannot be taught by demonstrations via kinesthetic teaching. Our approach can be easily executed on different humanoid robots. This removes the need for re-training, which is especially tedious when training with human-interaction partners. We show this by applying the learnt behaviours on two different humanoid robots with similar degrees of freedom but different shapes and control limits.
 
\end{abstract}

%%%%%%%%%%%%%%%%%%%%%%%%%%%%%%%%%%%%%%%%%%%%%%%%%%%%%%%%%%%%%%%%%%%%%%%%%%%%%%%%
\section{Introduction}

%According to the Computer-As-Social-Actor (CASA) paradigm, physical presence is an important predictor of the social response to robots in that humans put a robot in the same category as humans, also referred to as anthropomorphism \cite{nass1995anthropocentrism}. These are expected to be triggered by social human-like cues, such as voice \cite{nass1997machines}, face \cite{nass2001truth}, and language style \cite{nass1995anthropocentrism}. Users tend to look at computers as physiologically real \cite{nass2000machines}, by trying to apply embodied inherently human personality characteristics, to computers (and possibly robots by extension). 
% In the context of human-robot interaction (HRI), 
Physical contact, especially instantaneous contact is of great importance in various human-robot interaction (HRI) applications \cite{sakamoto2005cooperative}, especially since touch conveys information about the emotional state of a person \cite{hertenstein2006touch,yohanan2012role}. This enables a special kind of emotional connection to human users during the interaction \cite{han2012robotic,stock2018can}. Among such interactions, handshaking is a simple, natural interaction that is used in many social contexts \cite{chaplin2000handshaking,stewart2008exploring}. It plays an important role in shaping first impressions \cite{chaplin2000handshaking,stewart2008exploring} as it is usually the first non-verbal interaction that takes place in a social context. Thus, having a good handshake would not only widen the expressive abilities of a social robot but also provide a strong first impression for further interactions and is, therefore, an important skill required for the acceptance of social robots.

\begin{figure}
    \centering
    \includegraphics[width=0.4\textwidth]{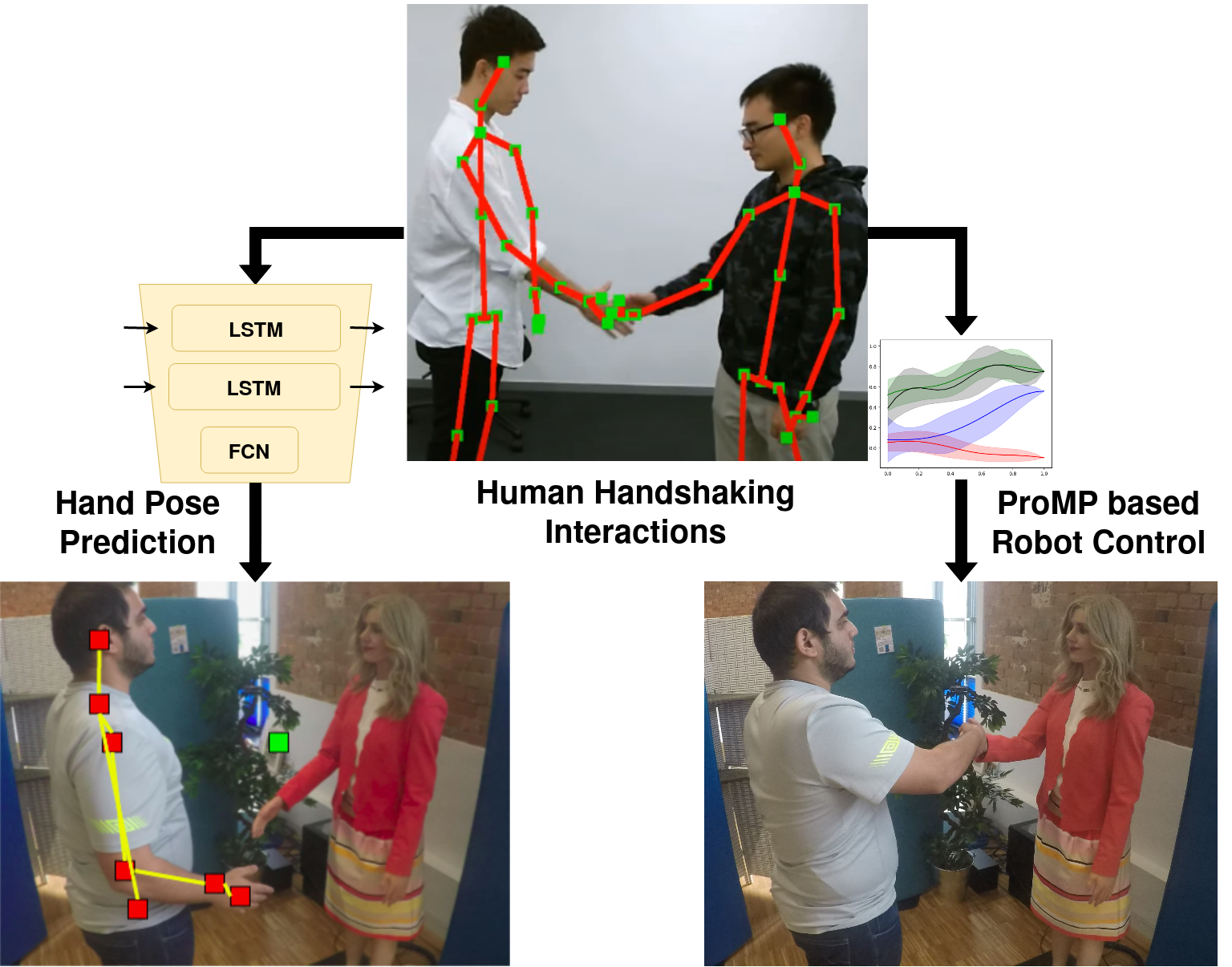}
    \caption{We propose a framework to learn human-robot handshaking from human-human interactions. Our training phase consists of two main steps. We first learn to predict the human's motions using a recurrent network and then learn a robot controller in the form of Probabilistic Movement Primitives (ProMPs) that can be conditioned in real-time using the predicted motions. Both are learnt just from skeleton data of human handshaking interactions, in a robot-agnostic fashion by leveraging the similarities between the degrees of freedom of humans and humanoid robots.}
    \label{fig:intro}
    % \vspace{-2em}
\end{figure}

The ``human-likeness" of robotic motions is an important aspect as movements can exaggerate feelings of uncanniness felt towards humanoid robots \cite{mori2012uncanny} as compared to their static appearance. Accordingly, the robot must be able to detect and predict the motion of the human body and react within a reasonable time. Otherwise, the physical interaction would be slow and unnatural. Vinayavekhin et al. \cite{vinayavekhin2017human} aim to bridge this gap in the context of human-robot handshaking using a recurrent network for predicting the human-hand motion and devise a simple controller for the robot's response motion. Such interaction dynamics during human-robot handshaking are also captured implicitly by Campbell et al. \cite{campbell2019learning} who learn a joint distribution over the trajectories of the human and the robot. However, their approach is robot-specific and would need to be re-trained with human interaction partners when applied to new robots. This can be more tedious than kinesthetic teaching for traditional robotic tasks. To some extent, this can be circumvented by learning from end-effector trajectories instead, like in \cite{christen2019guided}. They use Deep Reinforcement Learning with a human imitation reward to learn suitable motions but do not look at the interactive nature of the task. A detailed survey on human-robot handshaking can be found in \cite{prasad2020advances,prasad2021human}. 

To make way for a more adaptable method, we propose a framework for learning reaching behaviours for handshaking that can be easily transferred across different humanoid robots without re-training. We build on the framework proposed in \cite{vinayavekhin2017human} by improving the robot controller to learn from human-human interactions. This can be seen in Fig. \ref{fig:intro}. Moreover, in the case of pneumatically controlled robots which are not backdrivable and hence cannot be kinesthetically taught it is imperative to explore new ways of learning robot motions. 

Jindai and Watanabe \cite{jindai2007development} observed that hand motions of interaction partners are very similar in human-human handshaking. %Their data shows curved trajectories, similar to those shown in Fig. \ref{fig:hand_trajs}. 
Instead, Vinayavekhin et al. \cite{vinayavekhin2017human} propose simple straight-line trajectories to the detected hand location. In this regard, we learn directly from the joint motions of human-human interactions, similar to teleoperating a robot \cite{fritsche2015first}. We learn a distribution over the extracted joint angle trajectories in the form of Probabilistic Motion Primitives (ProMPs) \cite{paraschos2013probabilistic}. We exploit an important property of ProMPs which is that they can be conditioned to reach a particular location in their task space \cite{gomez2020adaptation}. We follow the recurrent motion prediction proposed in \cite{vinayavekhin2017human} to estimate the human hand's final location and condition the ProMPs to meet the human hand at the grasping point. We do so in a way that can seamlessly be executed on different humanoid/android robots, that have similar degrees of freedom as a human. Our main contribution is a principled pipeline that learns interactive actions directly from human demonstrations, removing the need to train robots with human interaction partners for such tasks. Since we learn the human joint motions, the learned behaviour can be easily transferred across different humanoid robots that have similar degrees of freedom. We show this by applying the learned behaviour on two humanoid robots, both of which are different in sizes and control limits.

\section{Preliminaries}
In this section, we provide a brief introduction of %the methods which we use in our approach, namely 
LSTMs in Sec. \ref{ssec:lstm} and ProMPs in Sec. \ref{ssec:promp}.
\subsection{Long Short Term Memory}
\label{ssec:lstm}
Long Short Term Memory (LSTM) networks \cite{hochreiter1997lstm} are a special kind of recurrent neural network (RNN) architecture for learning long term dependencies in time sequences. Typical RNNs just propagate the information forward from the previous timestep and combine it with the current timestep to get a prediction. LSTMs have multiple gates that decide how much information from previous time steps needs to be retained and how much from the current input would be used for the prediction. This allows them to learn long term correlations in temporal sequences, which makes them an ideal candidate for human motion prediction (a survey can be found at \cite{rasouli2020deep}). Moreover, we use the entire upper body 3D joints instead of just the arm or hand since body language can help in gauging the emotional state of a person \cite{sapinski2019emotion,metallinou2013tracking}. LSTMs have also shown good performance in this regard \cite{metallinou2013tracking}, which could be integrated into our approach in the future to predict emotional cues for handshaking as well.

\subsection{Probabilistic Movement Primitives}
\label{ssec:promp}
Probabilistic Movement Primitives (ProMPs)\cite{paraschos2013probabilistic} are a framework for learning robot trajectory distributions, where each trajectory is a sequence of $T$ observations $\boldsymbol{\tau} = [ \mathbf{y}_1 \dots \mathbf{y}_T]$, and each $\mathbf{y}_t$ is a vector of joint angles, velocities etc. at a time $t$. The trajectory distribution is parametrized as:
% \begin{equation}
%     \mathbf{y}_t = \begin{bmatrix}
%     \mathbf{\Phi}(t) & \dots & \mathbf{0} \\
%     \vdots& \ddots & \vdots \\
%     \mathbf{0} & \dots & \mathbf{\Phi}(t)
%   \end{bmatrix} \mathbf{\omega} + \mathbf{\epsilon}_\mathbf{y}
%     \label{eq:yt_traj}
% \end{equation}
% $\mathbf{y}_t = \mathbf{\Psi}(t)^T\boldsymbol{\omega} + \boldsymbol{\epsilon}_\mathbf{y}$
$p(\mathbf{y}_t|\boldsymbol{\omega}) = \mathcal{N}(\mathbf{y}_t|\mathbf{\Psi}(t)^T\boldsymbol{\omega}, \mathbf{\Sigma}_\mathbf{y})$
where $\mathbf{\Psi}(t)$ is a block diagonal matrix of time dependant basis functions, $\boldsymbol{\omega}$ is a weight vector and $\mathbf{\Sigma}_\mathbf{y}$, the covariance of the trajectories.
% This can be written as a probability distribution:

Variations in the trajectories are modelled by sampling $\boldsymbol{\omega}$ from a prior $\boldsymbol{\omega} \sim \mathcal{N}(\boldsymbol{\mu}_{\boldsymbol{\omega}}, \mathbf{\Sigma}_{\boldsymbol{\omega}})$. The likelihood can now be written in terms of the parameters $\boldsymbol{\theta}_{\boldsymbol{\omega}} = \{\boldsymbol{\mu}_{\boldsymbol{\omega}}, \mathbf{\Sigma}_{\boldsymbol{\omega}}\}$ as 
\begin{equation}
\label{eq:likelihood}
    p(\mathbf{y}_t;\boldsymbol{\theta}_{\boldsymbol{\omega}}) = \int \mathcal{N}(\mathbf{y}_t|\mathbf{\Psi}(t)^T\boldsymbol{\omega}, \mathbf{\Sigma}_\mathbf{y})\mathcal{N}(\boldsymbol{\omega}|\boldsymbol{\mu}_{\boldsymbol{\omega}}, \mathbf{\Sigma}_{\boldsymbol{\omega}})d\boldsymbol{\omega}
\end{equation}
The parameters $\boldsymbol{\mu}_{\boldsymbol{\omega}}, \mathbf{\Sigma}_{\boldsymbol{\omega}}$ are learnt by optimizing Eq. \ref{eq:likelihood} using maximum likelihood estimation over the demonstrations. 
% In our work we estimate a least-squares solution for $\mathbf{\omega}$ as $\mathbf{\omega}_i = argmin_{\mathbf{\omega}}  \rVert\mathbf{\Psi}^T\mathbf{\omega} - \mathbf{\tau}_i\rVert_2^2$ where $\mathbf{\tau}_i$ is the $i^{th}$ trajectory demonstration and  $\mathbf{\Psi}$ is the corresponding list of basis functions matrices. 
% $\mathbf{\mu}_\mathbf{\omega} = \frac{1}{N}\sum_{i=0}^N\mathbf{\omega}_i$ and \\
% $\mathbf{\Sigma}_\mathbf{\omega} = \frac{1}{N}\sum_{i=0}^N(\mathbf{\omega}_i - \mathbf{\mu}_\mathbf{\omega})(\mathbf{\omega}_i - \mathbf{\mu}_\mathbf{\omega})^T$

To counter the different speeds of executions in the demonstrations, a phase variable is used $z_t = \frac{t-t_0}{T}$ which is a normalized version of the actual trajectory time. This allows us to control the speed of execution while testing as well.

One important characteristic of ProMPs, that we exploit in this work, is the ability to condition them with particular observation(s) and applying Bayes rule to optimize the likelihood of the given observation(s). This can be done in two ways: joint space conditioning and task space conditioning.

In joint space conditioning, we wish to reach a given joint angle configuration $\mathbf{y}_t^*$ at time $t$. This is done by applying Bayes theorem at the given time with the observation $\{\mathbf{y}_t^*, \mathbf{\Sigma}_y^*\}$:
\begin{equation}
    p(\boldsymbol{\omega}|\mathbf{y}_t^*, \mathbf{\Sigma}_y^*) \propto \mathcal{N}(\mathbf{y}_t^*|\mathbf{\Psi}(t)^T\boldsymbol{\omega},\mathbf{\Sigma}_y^*)p(\boldsymbol{\omega})
\end{equation}

which yields a new distribution for $\boldsymbol{\omega}$ defined as:
\begin{equation}
\label{eq:mu_update}
    \boldsymbol{\mu}_{\boldsymbol{\omega}}^* = \boldsymbol{\mu}_{\boldsymbol{\omega}} + \mathbf{K}(\mathbf{y}_t^*-\mathbf{\Psi}(t)^T\boldsymbol{\mu}_{\boldsymbol{\omega}})
\end{equation}
\begin{equation}
\label{eq:sigma_update}
    \mathbf{\Sigma}_{\boldsymbol{\omega}}^* = \mathbf{\Sigma}_{\boldsymbol{\omega}} - \mathbf{K}\mathbf{\Psi}(t)^T\mathbf{\Sigma}_{\boldsymbol{\omega}}
\end{equation}
\begin{equation}
\label{eq:gain}
    \mathbf{K} = \mathbf{\Sigma}_{\boldsymbol{\omega}}\mathbf{\Psi}(t)(\mathbf{\Sigma}_y^* + \mathbf{\Psi}(t)^T\mathbf{\Sigma}_{\boldsymbol{\omega}}\mathbf{\Psi}(t))^{-1}
\end{equation}
In task space conditioning, given a target 3D location $\mathbf{x}_t^* \sim \mathcal{N}(\boldsymbol{\mu}_x, \mathbf{\Sigma}_x)$ at time $t$, a joint space configuration $\mathbf{y}_t$ is estimated such that it maximises $p(\mathbf{y}_t|\mathbf{x}_t^*, \boldsymbol{\theta}_{\boldsymbol{\omega}})$ to reach $\mathbf{x}_t$ while staying close to the learnt ProMP $\boldsymbol{\theta}_{\boldsymbol{\omega}}$. Further details about task-space conditioning can be found in \cite{gomez2020adaptation}.

\begin{figure*}
\vspace{2em}
    \centering
    \includegraphics[width=0.9\textwidth]{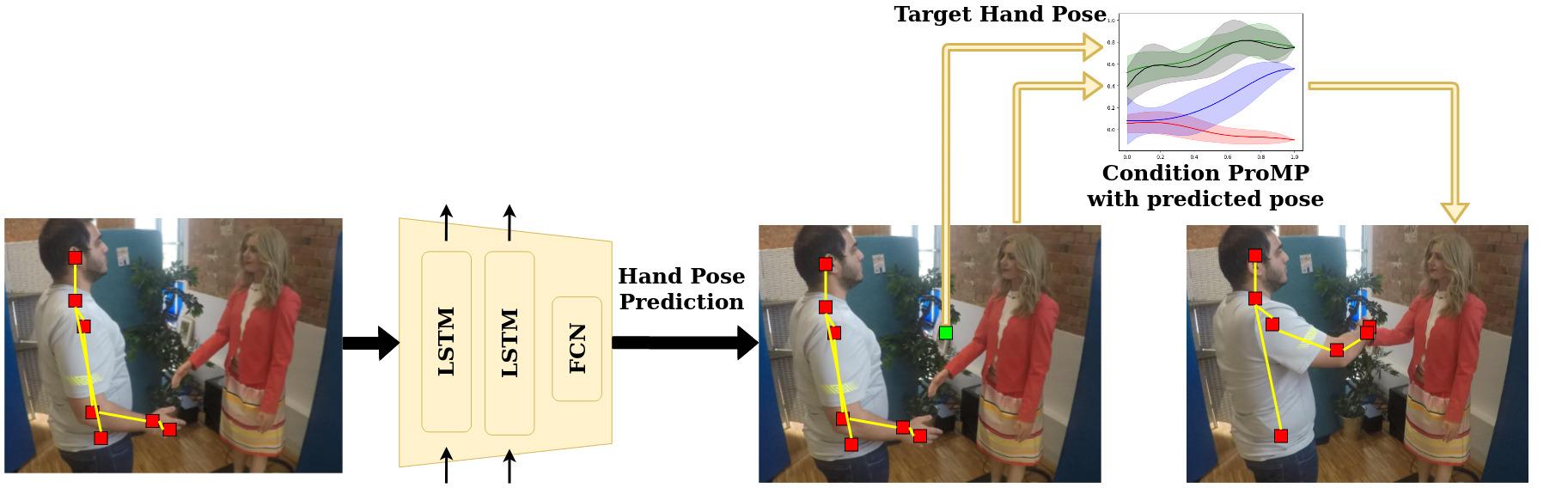}
    \caption{Overview of our proposed method. We extract the human skeleton which is given to a recurrent network to predict the final hand location. The predictions are used, at each timestep, to condition the learnt ProMP controller so that the robot hand meets the human hand at the final location.}
    \label{fig:flow}
    % \vspace{-2em}
\end{figure*}

\section{Proposed Approach}
In this section, we provide a detailed explanation of our novel approach of learning robot-agnostic hand reaching behaviours from human interactions using recurrent networks for explicitly encoding the interaction dynamics and using ProMPs for learning robot control from human skeleton data. This is especially useful as one of the robots is non-backdrivable and cannot be taught via kinesthetic teaching. An overview of this novel framework can be seen in Fig. \ref{fig:flow} and is explained in Sec. \ref{ssec:overview}. The hand location prediction for enabling the ``interactiveness" of the method is explained in Sec. \ref{ssec:lstm_learning}. Learning ProMPs from the human skeleton data for humanoid robots is explained in Sec. \ref{ssec:promp_learning}. The combination of the hand prediction and the subsequent conditioning of the ProMP with the predicted location is explained in Sec. \ref{ssec:promp_hand_integration}.

\subsection{Overview}
\label{ssec:overview}
Given skeleton data from human-human interactions, we use the 3D upper body joint locations to train the recurrent network to predict the final hand location. Learning the ProMP can be thought of as teaching the robot via teleoperation. We extract joint angles from the skeleton data which are similar as the degrees of freedom of the humanoid robots (shoulder yaw, pitch, roll and elbow angle) with which we learn the ProMP over the extracted joint angle trajectories. This circumvents the need for kinesthetic teaching which is not possible for non-backdrivable robots like one of the robots used in this work. During testing, the human interaction partner's skeleton is tracked by an external RGB-D camera. With this, the network predicts a final hand location. Given the predicted and the true hand location, a target hand location is formulated and the ProMP is conditioned to meet the hand at the target location. This is done to smoothly shift from the predicted location to the true location as we humans do during such an interaction.
\subsection{Predicting Human Hand Motions using LSTMs}
\label{ssec:lstm_learning}
During handshaking and other interactive behaviours, we as humans try to predict where the interaction partner's hand would go based on the motion and body language and move accordingly, adapting the motion to meet the hand at the final location as the interaction progresses. To explicitly encode this kind of predictive behaviour, we train an LSTM using the upper body skeleton joint locations from the human-human interaction sequences. We use the same architecture as in \cite{vinayavekhin2017human} where given a sequence of $n$ frames of upper body skeleton joint locations in 3D $\mathbf{b}_1, \mathbf{b}_2, \mathbf{b}_3 \dots \mathbf{b}_n (  \mathbf{b}_i \in \mathbb{R}^{15\times3})$, the final 3D hand location $\mathbf{\hat{h}}_n$ is predicted at each time step. The upper body is used since it allows us to learn relations between body language and the motion prediction, instead of using only the hand or arm motions.  %The network consists of 2 LSTM layers with a hidden dimension of 64 followed by a fully connected layer that predicts the final 3D hand location at each time-step.
We prefer this approach over learning the joint trajectories of both the human and the robot since this can be explicitly used for any robot that the human interacts with, rather than learning robot-specific interactions, which may not be as easy to transfer to a different robot.

\subsection{Learning from Human Motions using ProMPs}
\label{ssec:promp_learning}
% \textbf{TODO: IMAGES OF HUMAN SKELETONS WITH ANGLES AND CORRESPONDING ROBOT POSITIONS. (IDEA: 3x3 Table with first row showing shoulder yaw, 2nd for shoulder pitch and third for shoulder roll. The first column is human, the second column is pepper, third is Elenoide)}

\begin{figure}
    \centering
    % \hfill
    \begin{subfigure}[b]{0.14\textwidth}
        \centering
        \includegraphics[width=\textwidth]{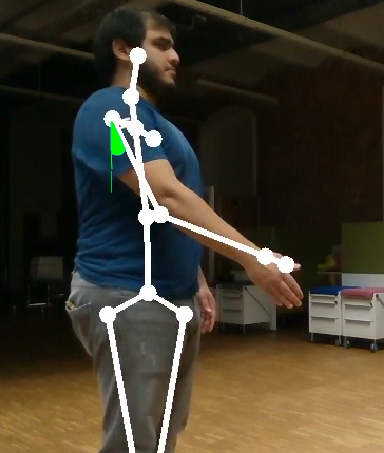}
        % \caption{Experimental setup with Pepper \cite{pandey2018mass}.}
        % \label{fig:pepper}
    \end{subfigure}%
    % \hfill
    \begin{subfigure}[b]{0.165\textwidth}
        \centering
        \includegraphics[width=\textwidth]{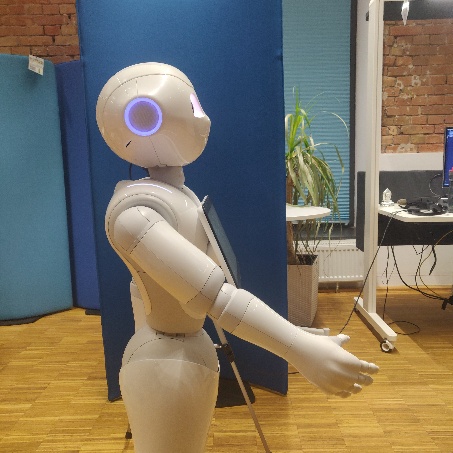}
        % \caption{Experimental setup with Pepper \cite{pandey2018mass}.}
        % \label{fig:pepper}
    \end{subfigure}%
    % \hfill
    \begin{subfigure}[b]{0.165\textwidth}
        \centering
        \includegraphics[width=\textwidth]{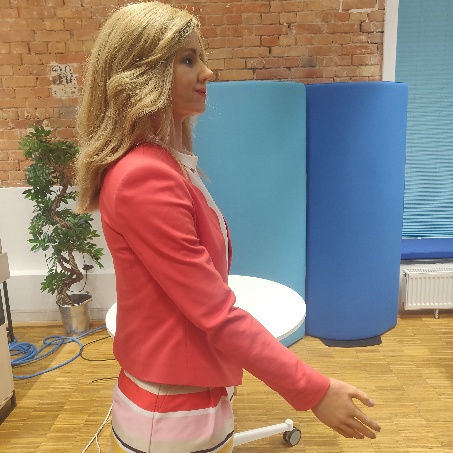}
        % \caption{The pneumatic robot, Elenoide, used in this work.}
        % \label{fig:elenoide}
    \end{subfigure}
    % \hfill
    
    % \hfill
    \begin{subfigure}[b]{0.14\textwidth}
        \centering
        \includegraphics[width=\textwidth]{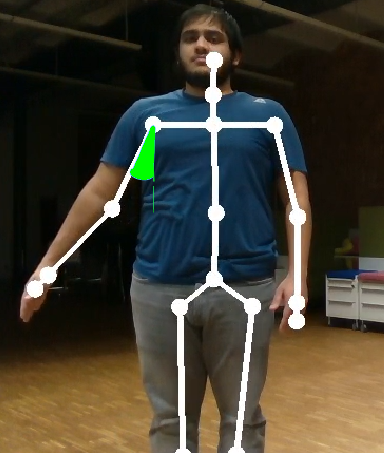}
        % \caption{Experimental setup with Pepper \cite{pandey2018mass}.}
        % \label{fig:pepper}
    \end{subfigure}%
    % \hfill
    \begin{subfigure}[b]{0.165\textwidth}
        \centering
        \includegraphics[width=\textwidth]{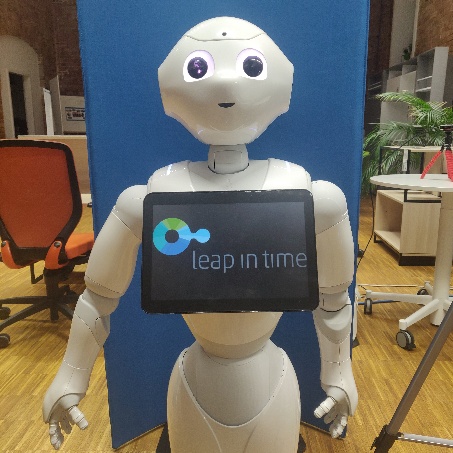}
        % \caption{Experimental setup with Pepper \cite{pandey2018mass}.}
        % \label{fig:pepper}
    \end{subfigure}%
    % \hfill
    \begin{subfigure}[b]{0.165\textwidth}
        \centering
        \includegraphics[width=\textwidth]{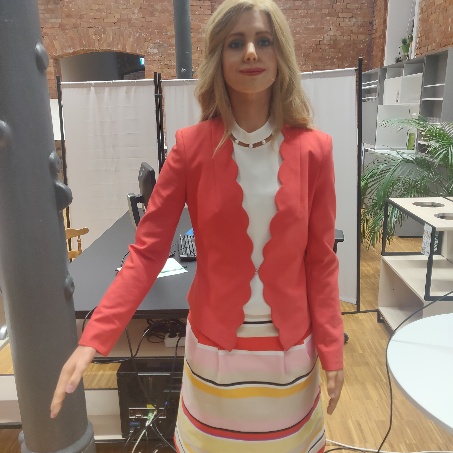}
        % \caption{The pneumatic robot, Elenoide, used in this work.}
        % \label{fig:elenoide}
    \end{subfigure}
    % \hfill
    
    % \hfill
    \begin{subfigure}[b]{0.14\textwidth}
        \centering
        \includegraphics[width=\textwidth]{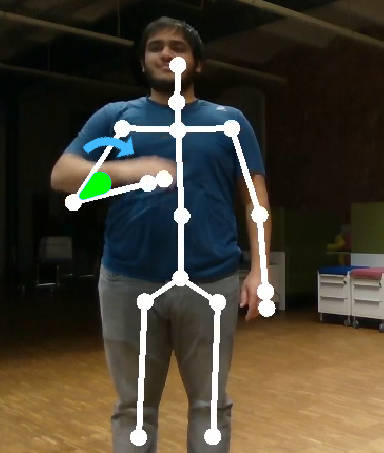}
        % \caption{Experimental setup with Pepper \cite{pandey2018mass}.}
        % \label{fig:pepper}
    \end{subfigure}%
    % \hfill
    \begin{subfigure}[b]{0.165\textwidth}
        \centering
        \includegraphics[width=\textwidth]{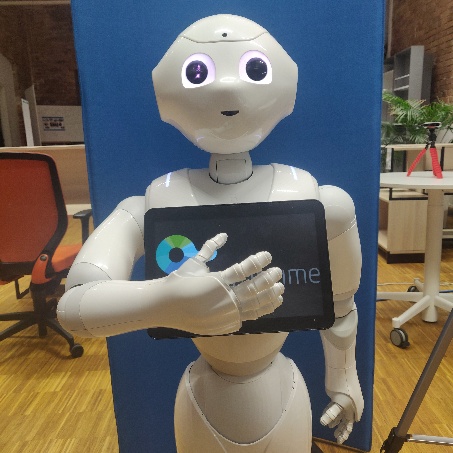}
        % \caption{Experimental setup with Pepper \cite{pandey2018mass}.}
        % \label{fig:pepper}
    \end{subfigure}%
    % \hfill
    \begin{subfigure}[b]{0.165\textwidth}
        \centering
        \includegraphics[width=\textwidth]{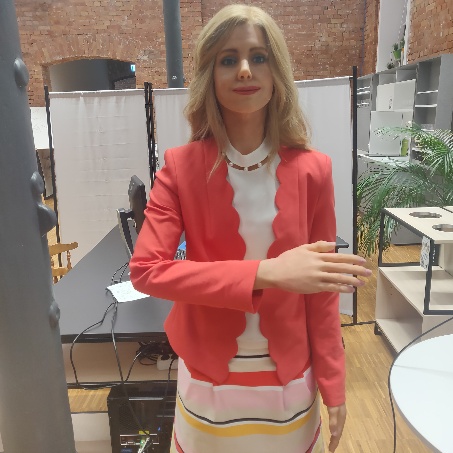}
        % \caption{The pneumatic robot, Elenoide, used in this work.}
        % \label{fig:elenoide}
    \end{subfigure}
    % \hfill
    \caption{Joint angle extraction from 3D skeleton data. The shoulder and elbow angles (marked in blue in the first column) are extracted using the skeleton geometry to directly control the robots, like in a teleoperation scenario.}
    \label{fig:teleoperation}
    % \vspace{-2em}
\end{figure}
Given the 3D skeleton data of a human, the shoulder angles (roll, pitch, yaw) and the elbow angle are extracted for the right hand (details can be found in \cite{fritsche2015first}), which can be seen as learning by teleoperation. We use the joint angles instead of the 3D joint locations as the angles capture the underlying control that we as humans perform and more importantly, this is similar to the degrees of freedom present for the humanoid robots that we use in this work, as shown in Fig. \ref{fig:teleoperation}. Since the wrist angles are difficult to compute due to errors in the tracking, we defer this to our future work. While using the joint angles would lead to different end-effector locations based on the robot kinematics, they can be additionally fine-tuned using inverse kinematics to reach a given behaviour, as explained in Sec. \ref{ssec:promp_hand_integration}.

Given the extracted joint angle trajectories, %we fit a ProMP to learn a distribution for the 4 degrees of freedom in question. 
the ProMP weights are computed by optimizing Eq. \ref{eq:likelihood} using ridge regression as $\mathbf{\omega}_i = (\mathbf{\Psi}^T\mathbf{\Psi} + \lambda \mathbf{I})^{-1}\mathbf{\Psi}^T\boldsymbol{\tau}_i$ where $\boldsymbol{\tau}_i$ is the $i^{th}$ trajectory and $\mathbf{\Psi}$ is the corresponding list of basis functions matrices. We set $\lambda = 10^{-10}$ as larger values cause a divergence in the learning\cite{paraschos2018using}. Additionally, as shown in \cite{paraschos2018using}, the jerk can be minimized as $\boldsymbol{\omega}_i = (\mathbf{\Psi}^T\mathbf{\Psi} + \lambda \mathbf{\Gamma}^T\mathbf{\Gamma})^{-1}\mathbf{\Psi}^T\boldsymbol{\tau}_i$ where $\mathbf{\Gamma}$ is the third derivative of $\mathbf{\Psi}$ w.r.t time. From this, we can calculate the ProMP parameters $\boldsymbol{\mu}_{\boldsymbol{\omega}}, \mathbf{\Sigma}_{\boldsymbol{\omega}}$ as the mean and covariance of the estimated weight vectors.

%  Moreover, learning joint locations would additionally require an understanding of the dynamics between the different joints itself. Currently, we do not use the second human's skeletons in the frame for learning the underlying interaction dynamics. Instead, we impose a more explicit interaction modelling using the predicted hand pose since ProMPs can be conditioned to reach a given 3D point.

\subsection{ProMP Conditioning with Predicted Hand Locations}
\label{ssec:promp_hand_integration}

During testing, the human's skeleton is tracked and the upper body joints are given to the LSTM network to predict a final hand location. Like in \cite{vinayavekhin2017human}, to shift to the human hand towards the end of the trajectory, thereby converging to the true hand location, the target location at the give time step $\mathbf{h}^*_t$ for conditioning the ProMP is calculated as:
\begin{equation}
    \mathbf{h}^*_t = (1 - \sigma(t + \alpha))\mathbf{\hat{h}}_t + \sigma(t + \alpha)\mathbf{h}_t
\label{eq:target_location}
\end{equation}
where $\mathbf{\hat{h}}_t$ is the predicted hand location from the LSTM, $\mathbf{h}_t$ is the tracked hand location, $\sigma(\cdot)$ is the sigmoid function and $\alpha$ is constant to center the sigmoid at half the trajectory. We set $\alpha = 0.67$, which corresponds to approximately 20 frames, which is the obtained from the average of the training trajectories. %A pipeline of the testing phase can be seen in Fig. \ref{fig:robot_testing}.
This ensures a smooth transition between the predicted and tracked hand location towards the end of the interaction. 

Given the target hand location, the ProMP is conditioned by maximising $p(\mathbf{y}_t|\mathbf{x}_t^*, \boldsymbol{\theta}_{\boldsymbol{\omega}})$ where $\mathbf{x}_t^* = \mathbf{h}^*_t$. This boils down to the following optimization problem\cite{gomez2020adaptation}: 
\begin{equation}
    % \mathbf{y}_t^* = \argminC_\mathbf{y}  \lambda_x\rVert\boldsymbol{\mu}_\mathbf{x} - f(\mathbf{y})\lVert_{\mathbf{\Sigma}_\mathbf{x}}^2 + \lambda_y\rVert\boldsymbol{\mu}_\mathbf{y} - \mathbf{y}\lVert_{\mathbf{\Sigma}_\mathbf{y}}^2
    \mathbf{y}_t^* = \argmaxC_\mathbf{y} \mathcal{N}(f(\mathbf{y})|\mathbf{x}_t^*, \mathbf{\Sigma}_\mathbf{x}) \mathcal{N}(\mathbf{y}|\boldsymbol{\mu}_{\mathbf{y}(t)}, \mathbf{\Sigma}_{\mathbf{y}}(t)) 
    \label{eq:inv_kin}
\end{equation}
where $f(\mathbf{y})$ is the forward kinematics to estimate the end-effector location given joint angles $\mathbf{y}$, $\boldsymbol{\mu}_\mathbf{y} = \mathbf{\Psi}(t)^T\boldsymbol{\mu}_{\boldsymbol{\omega}}$ and $\mathbf{\Sigma}_\mathbf{y} = \mathbf{\Psi}(t)^T\mathbf{\Sigma}_{\boldsymbol{\omega}}\mathbf{\Psi}(t)$ are the marginal distribution of the learnt ProMP and $\mathbf{\Sigma}_{\mathbf{x}}$ is the desired accuracy in the task space. Given $\mathbf{y}_t^*$, we perform joint space conditioning to obtain the new ProMP parameters (Eqs. \ref{eq:gain} - \ref{eq:sigma_update})
\begin{figure}[h!]
% \vspace{-1em}centering
    \hfill
    \begin{subfigure}[b]{0.2\textwidth}
        \centering
        \includegraphics[width=\textwidth]{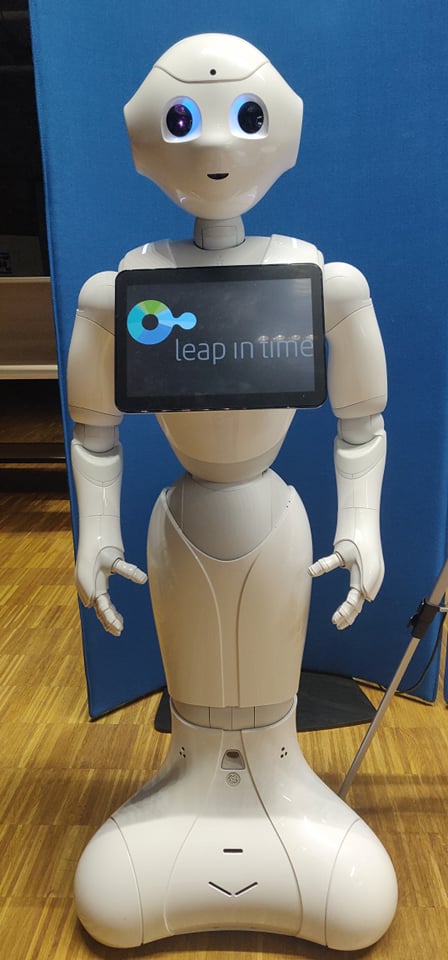}
        \caption{Pepper}
        \label{fig:pepper}
    \end{subfigure}%
    \hfill
    \begin{subfigure}[b]{0.2\textwidth}
        \centering
        \includegraphics[width=\textwidth]{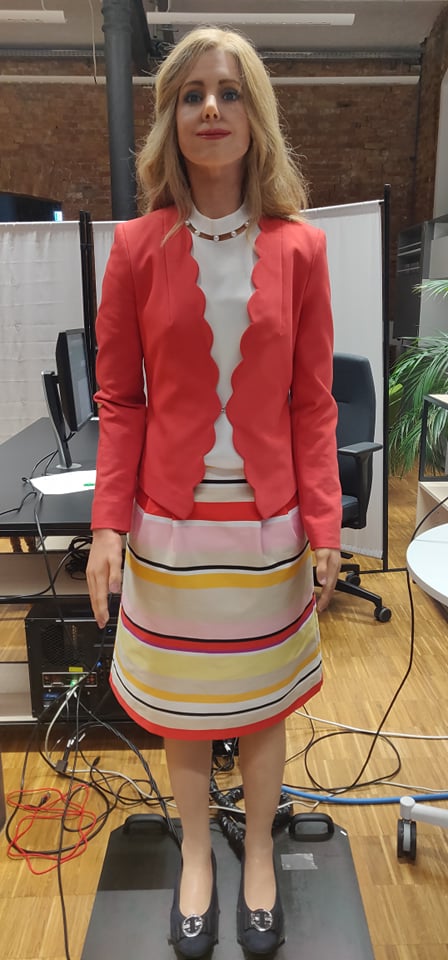}
        \caption{Elenoide}
        \label{fig:elenoide}
    \end{subfigure}
    \hfill
\caption{Humanoid robots used in this work}
\label{fig:robots}
% \vspace{-1em}
\end{figure}
\section{Experiments and Results}
In this section, we introduce the robots and go into further detail about the implementation of our method (Sec. \ref{ssec:setup}), the dataset used to train the human motion prediction network and the robot-agnostic reaching behaviour (Sec. \ref{ssec:dataset}) and the results of the LSTM network predictions and reaching behaviours on different robots (Sec. \ref{ssec:results}). 
\subsection{Experimental Setup}
\label{ssec:setup}
We use two humanoid robots in our experiments. The first is Pepper (Fig. \ref{fig:pepper}), a humanoid social robot from Aldebaran Robotics \cite{pandey2018mass}, having 6 degrees of freedom in each arm. The second is a custom-made pneumatically controlled android robot, Elenoide (Fig. \ref{fig:elenoide}), having 14 degrees of freedom per arm. We use NuiTrack\cite{nuitrack} on data captured by an Intel RealSense \cite{keselman2017intel} to track the human partner's skeleton. The entire system is controlled using ROS \cite{quigley2009ros}.

The hand prediction network is implemented using PyTorch \cite{paszke2019pytorch} and consists of 2 LSTM layers with a hidden dimension of 64 followed by a Fully Connected Netowrk (FCN) layer that predicts the final 3D hand location at each time-step. The network was trained with a batch size of 32 for 200 epochs using the Adam optimizer \cite{kingma2014adam}. Given that Eq. \ref{eq:target_location} needs an estimate of the trajectory length, we use an estimated trajectory length of 32 frames ($\sim 1$ sec.), which is the median length of the training trajectories. For learning the ProMP, 3 RBF kernels with equally spaced centers and a scale of 0.01 are used. The ProMPs are implemented using a modified version\footnote{\url{https://github.com/souljaboy764/intprim}} of the Bayesian Interaction Primitives Framework \cite{campbell2017bayesian}. We use SciPy \cite{scipy} for the least-squares estimation of the ProMP weights and for solving the inverse kinematics during task space conditioning (Eq. \ref{eq:inv_kin}).%, with $\lambda_x = \lambda_y = 0.5$. 

\subsection{Dataset}
\label{ssec:dataset}
We use the skeleton data from the handshaking interactions present in the NTU RGB+D dataset \cite{Shahroudy_2016_NTURGBD} for training the hand prediction network and for learning the ProMPs. The skeleton data is recorded using a Microsoft Kinect \cite{zhang2012microsoft} v2, which provides as a set of 25 joints for each person present in the video, of which we use the 15 upper body joints.  %Fig. \ref{fig:hand_trajs} shows some sample hand trajectories of one of the interaction partners from the handshaking interactions. The darker shade denotes the starting of the trajectories and the green colour denotes the ending of the trajectories. 

% \begin{figure}[h!]
%     \centering
%     \includegraphics[width=0.3\textwidth]{images/hand_trajs.png}
%     \caption{Sample hand trajectories for handshaking from the NTU RGB+D dataset. (dark - starting, bright - ending)}
%     \label{fig:hand_trajs}
%     \vspace{-1em}
% \end{figure}

\begin{figure*}[ht!]
\vspace{2em}
    \centering
     \begin{subfigure}[b]{0.2\textwidth}
        \centering
        \includegraphics[width=\textwidth]{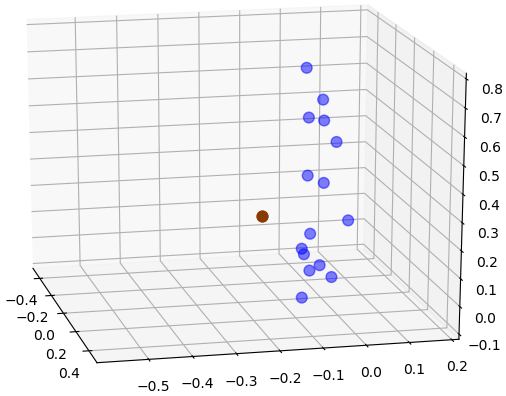}
        \caption{Starting frame}
        \label{fig:lstm_pred_0}
    \end{subfigure}%
     \begin{subfigure}[b]{0.2\textwidth}
        \centering
        \includegraphics[width=\textwidth]{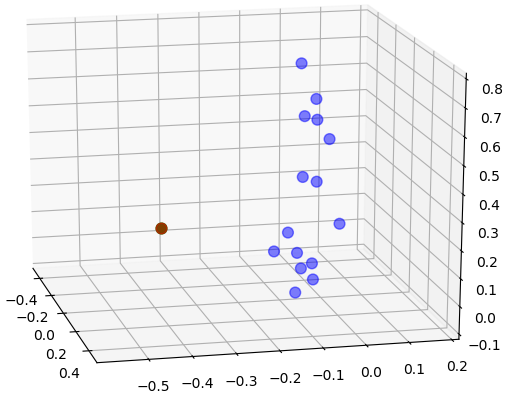}
        \caption{25\%}
        \label{fig:lstm_pred_5}
    \end{subfigure}%
     \begin{subfigure}[b]{0.2\textwidth}
        \centering
        \includegraphics[width=\textwidth]{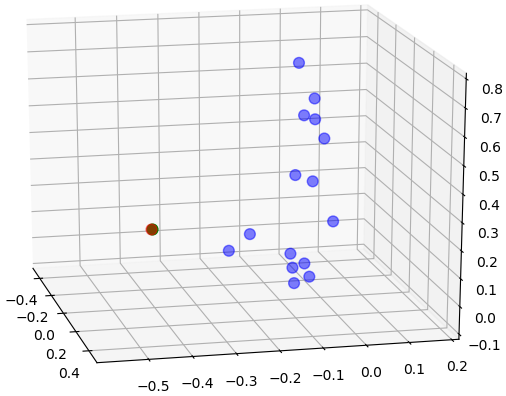}
        \caption{50\%}
        \label{fig:lstm_pred_10}
    \end{subfigure}%
     \begin{subfigure}[b]{0.2\textwidth}
        \centering
        \includegraphics[width=\textwidth]{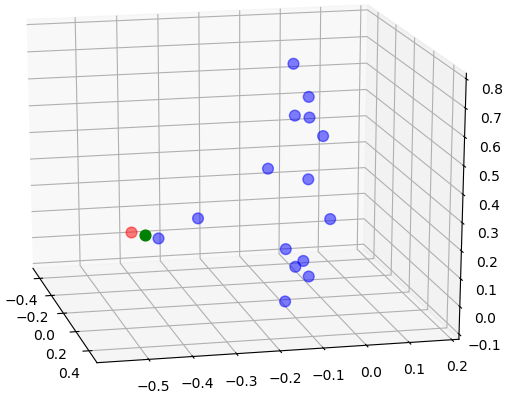}
        \caption{75\%}
        \label{fig:lstm_pred_15}
    \end{subfigure}%
     \begin{subfigure}[b]{0.2\textwidth}
        \centering
        \includegraphics[width=\textwidth]{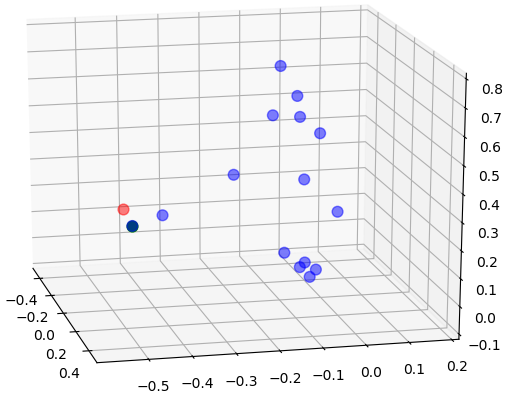}
        \caption{Final frame}
        \label{fig:lstm_pred_20}
    \end{subfigure}%

    \caption{Hand Location Prediction Example. The leftmost frame is the starting frame, the rightmost is the final frame, the others are equally spaced between them. The blue dots denote the upper body skeleton joints, the red dot is the predicted final location from the network and the green dot is the target location calculated using Eq. \ref{eq:target_location}.}%As it can be seen, our prediction is accurate within a few centimetres, which is permissible as the target location gradually shifts to the final hand location.}
    \label{fig:lstm_pred}
    % \vspace{-2em}
\end{figure*}

From the skeletons of the handshaking action, we remove those in which the handshaking is done with the left hand. We further remove those which have errors or discontinuities in the skeleton tracking. Among those remaining, we filter out the part of the trajectory after the hands are grasped and the initial part of the trajectory where no movement is present. Of these, we use 500 trajectories, randomly split into 400 training and 100 testing trajectories to train the hand prediction network. We further filter out trajectories that still have irregularities and use the remaining 197 trajectories and extract the joint angles of the right hand from each of the trajectories to train the ProMP. Further details about the joint angle extraction can be found in \cite{fritsche2015first}.

\subsection{Results}
\label{ssec:results}
\subsubsection{Hand Prediction Accuracy}
\begin{figure}[h!]
% \vspace{-2em}
    \centering
    \includegraphics[width=0.45\textwidth]{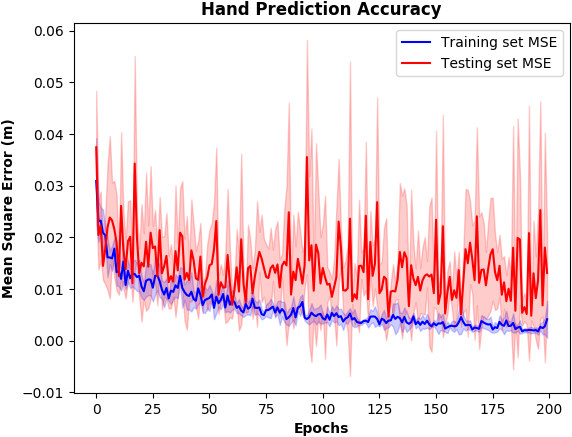}
    \caption{Accuracy of the hand prediction network.}
    \label{fig:hand_pred_acc}
    % \vspace{-1em}
\end{figure}

In Fig. \ref{fig:hand_pred_acc}, we show the training loss and testing loss as the training progresses. % using 50\% and 80\% of the trajectory. 
As it can be seen, the network predictions are accurate within a few centimetres, which can be seen in Fig. \ref{fig:lstm_pred}. This is not an issue since the prediction is only used in the initial stages of the interaction. The target location calculated using Eq. \ref{eq:target_location} switches smoothly to the true tracked location as the trajectory comes towards the end. %This is also shown in Fig. \ref{fig:hand_target_traj} where the target trajectory (green) can be seen shifting from the network prediction (red) to the true location (blue).
% \begin{figure}[h!]
%     \centering
%     \includegraphics[width=0.3\textwidth]{images/hand_target_traj.png}
%     \caption{Example of the 3D target location generation used to condition the ProMP. The target location (green) gradually shifts from the predicted location (red) to the actual hand trajectory (blue). The final prediction and the true hand location are plotted on the top left side.}
%     \label{fig:hand_target_traj}
%     \vspace{-2em}
% \end{figure}

\subsubsection{Hand Reaching Results}
\begin{figure*}[h!]
\vspace{2em}
\centering
    \begin{subfigure}[b]{0.24\textwidth}
    \includegraphics[width=\textwidth]{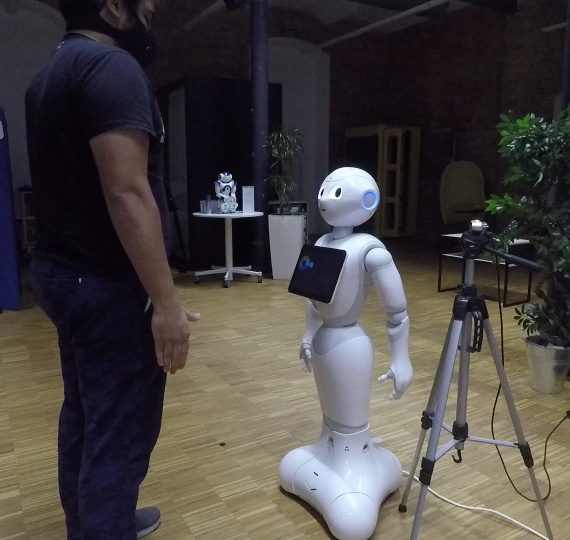}
    % \caption{}
    % \label{fig:sim_pose_0}
    \end{subfigure}%
    \begin{subfigure}[b]{0.24\textwidth}
    \includegraphics[width=\textwidth]{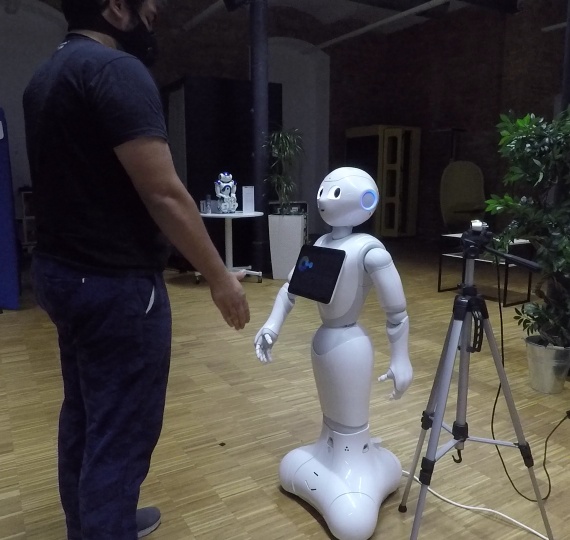}
    % \caption{}
    % \label{fig:sim_pose_25}
    \end{subfigure}%
    \begin{subfigure}[b]{0.24\textwidth}
    \includegraphics[width=\textwidth]{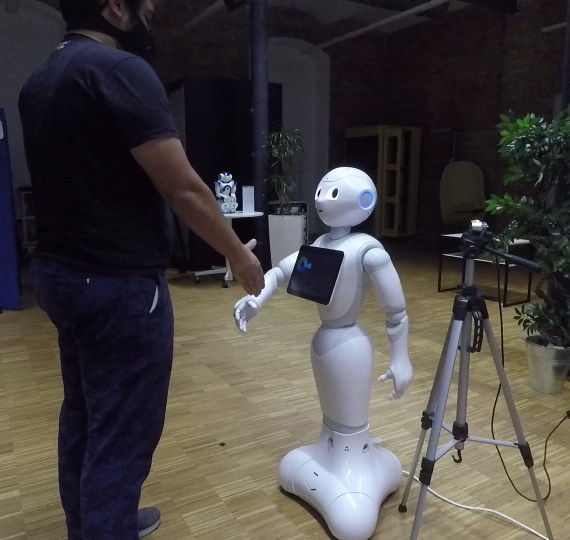}
    % \caption{}
    % \label{fig:sim_pose_50}
    \end{subfigure}%
    \begin{subfigure}[b]{0.24\textwidth}
    \includegraphics[width=\textwidth]{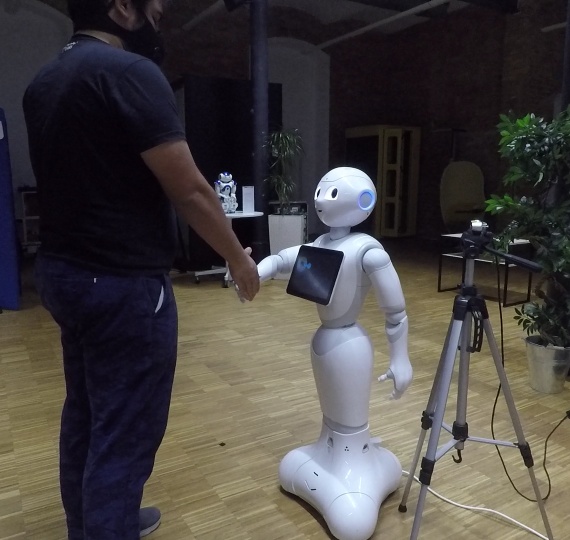}
    % \caption{}
    % \label{fig:sim_pose_75}
    \end{subfigure}%

\begin{subfigure}[b]{0.24\textwidth}
    \includegraphics[width=\textwidth]{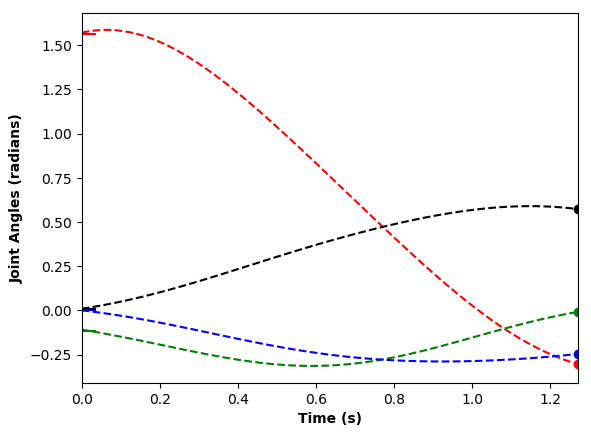}
    % \caption{}
    % \label{fig:sim_pose_0}
    \end{subfigure}%
    \begin{subfigure}[b]{0.24\textwidth}
    \includegraphics[width=\textwidth]{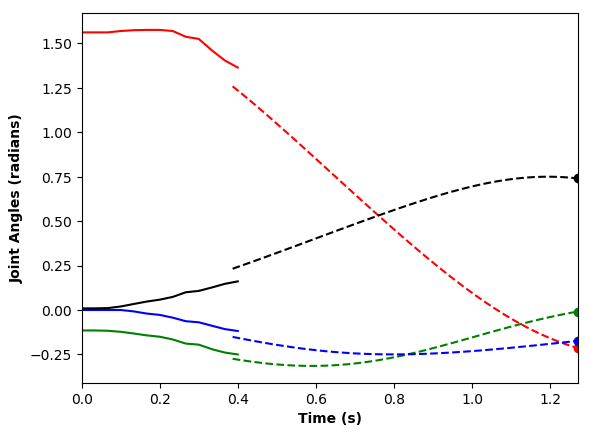}
    % \caption{}
    % \label{fig:sim_pose_25}
    \end{subfigure}%
    \begin{subfigure}[b]{0.24\textwidth}
    \includegraphics[width=\textwidth]{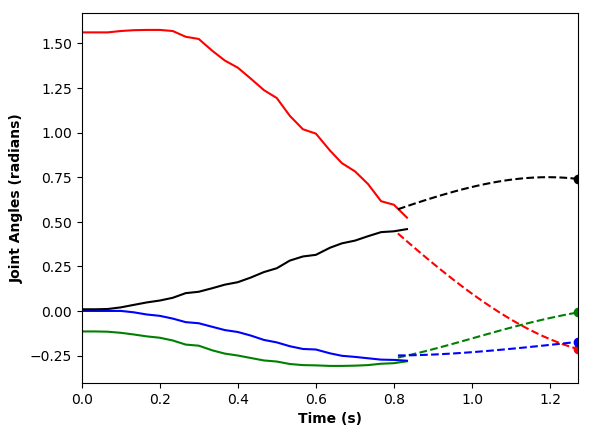}
    % \caption{}
    % \label{fig:sim_pose_50}
    \end{subfigure}%
    \begin{subfigure}[b]{0.24\textwidth}
    \includegraphics[width=\textwidth]{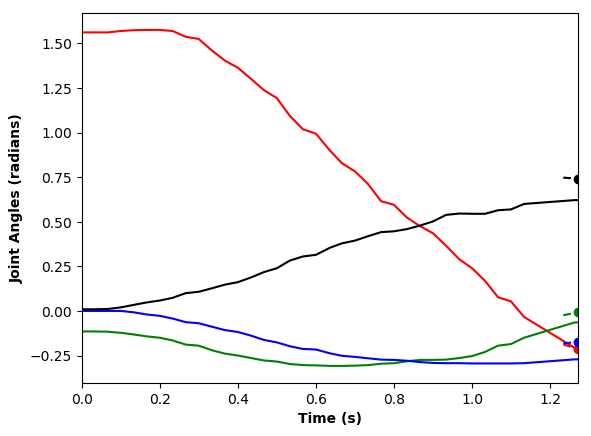}
    % \caption{}
    % \label{fig:sim_pose_75}
    \end{subfigure}%

\begin{subfigure}[b]{0.24\textwidth}
    \includegraphics[width=\textwidth]{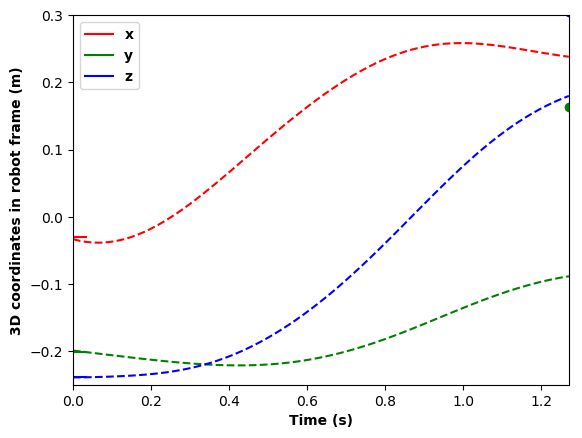}
    % \caption{}
    % \label{fig:sim_pose_0}
    \end{subfigure}%
    \begin{subfigure}[b]{0.24\textwidth}
    \includegraphics[width=\textwidth]{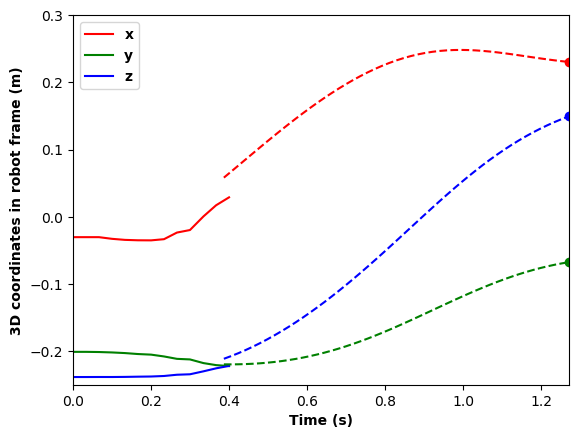}
    % \caption{}
    % \label{fig:sim_pose_25}
    \end{subfigure}%
    \begin{subfigure}[b]{0.24\textwidth}
    \includegraphics[width=\textwidth]{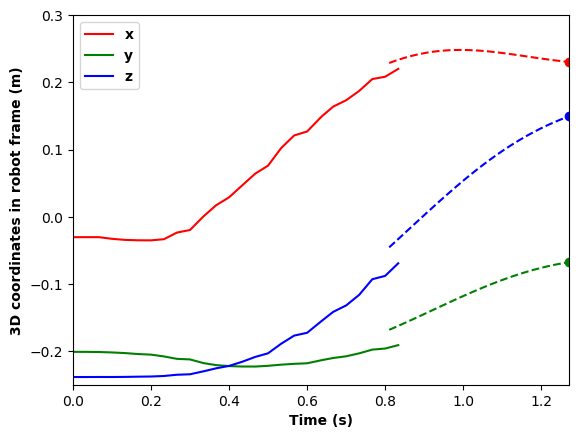}
    % \caption{}
    % \label{fig:sim_pose_50}
    \end{subfigure}%
    \begin{subfigure}[b]{0.24\textwidth}
    \includegraphics[width=\textwidth]{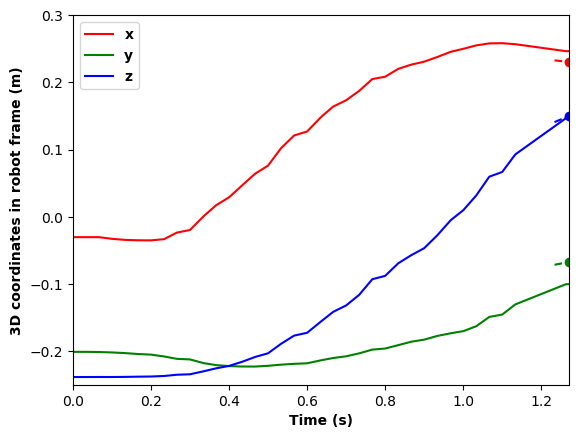}
    % \caption{}
    % \label{fig:sim_pose_75}
    \end{subfigure}%
 \caption{An example of an interaction generated by our proposed approach. The top row shows the interaction, the second row shows the robot joint angles and the third row shows the end effector location in the robot's frame. The solid lines denote the observed values, the dashed lines represent the values from the commands generated by the ProMP and the dot at the end represents the target value used to condition the ProMP.}
%  \vspace{-3em}
 \label{fig:interaction}
\end{figure*}

\begin{figure*}[h!]
\centering
    \begin{subfigure}[b]{0.24\textwidth}
    \includegraphics[width=\textwidth]{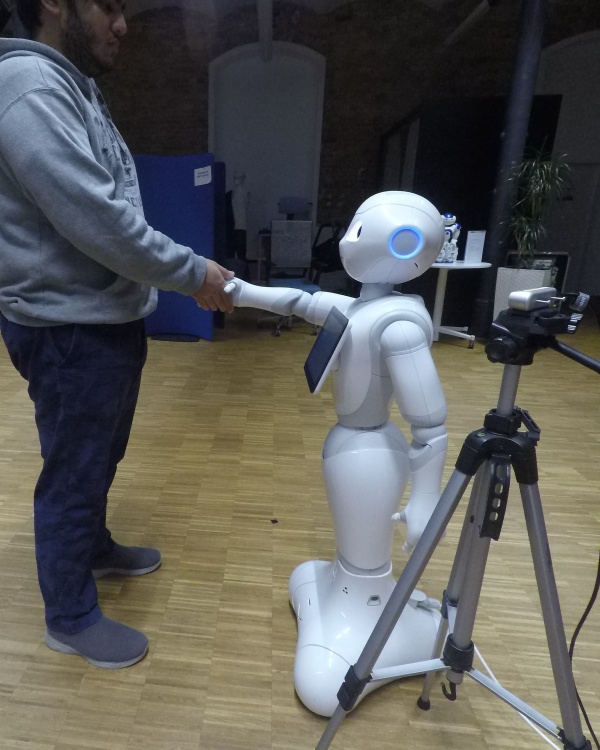}
    % \caption{}
    % \label{fig:sim_pose_0}
    \end{subfigure}%
    \begin{subfigure}[b]{0.24\textwidth}
    \includegraphics[width=\textwidth]{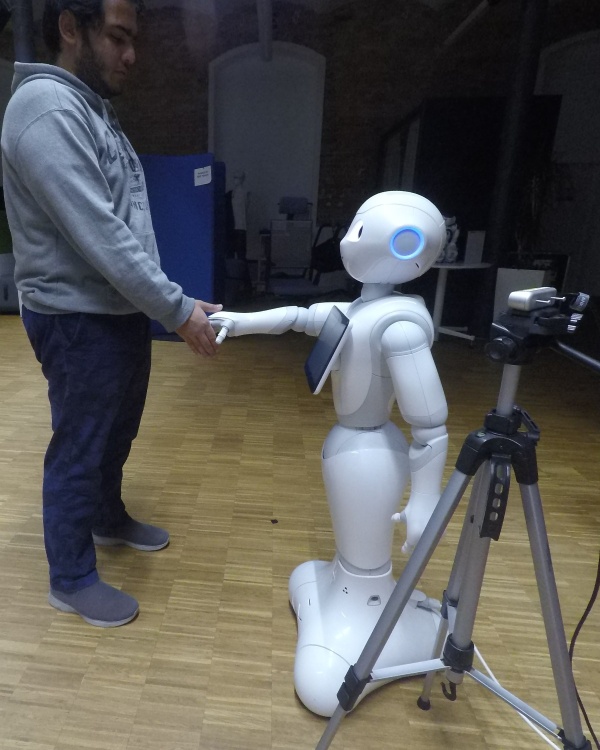}
    % \caption{}
    % \label{fig:sim_pose_25}
    \end{subfigure}%
    \begin{subfigure}[b]{0.24\textwidth}
    \includegraphics[width=\textwidth]{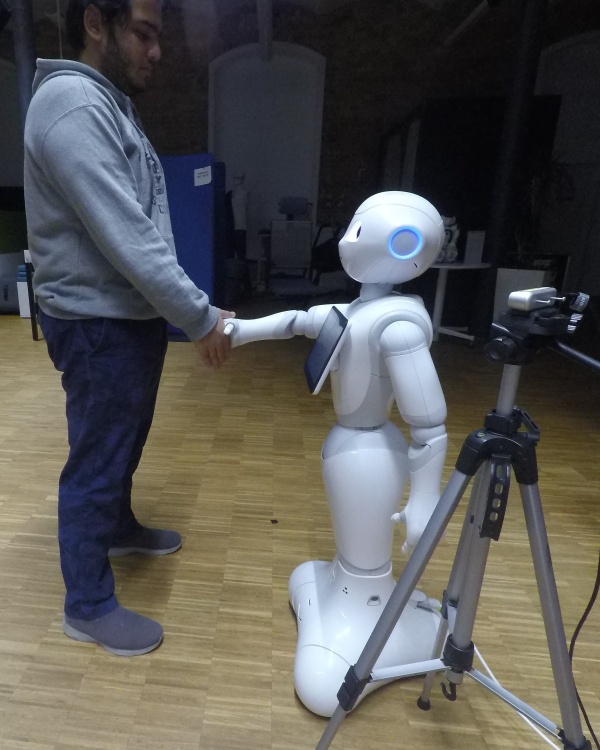}
    % \caption{}
    % \label{fig:sim_pose_50}
    \end{subfigure}%
    \begin{subfigure}[b]{0.24\textwidth}
    \includegraphics[width=\textwidth]{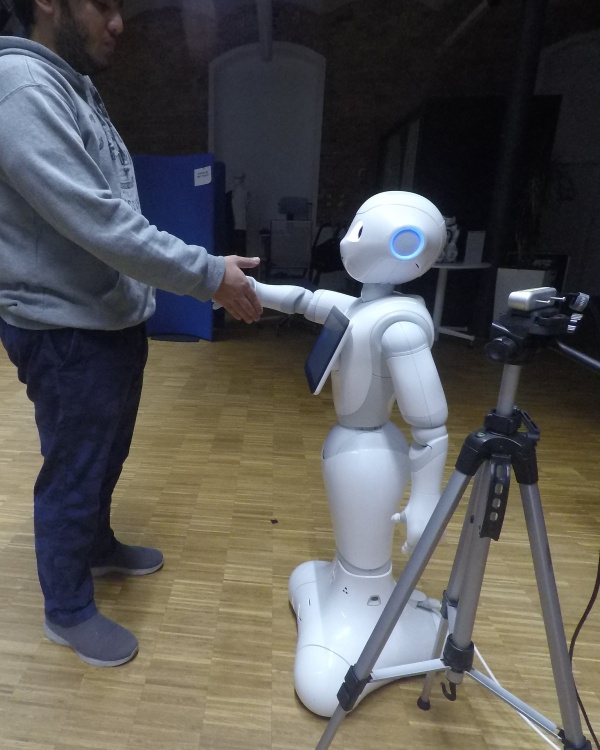}
    % \caption{}
    % \label{fig:sim_pose_75}
    \end{subfigure}%

\begin{subfigure}[b]{0.24\textwidth}
    \includegraphics[width=\textwidth]{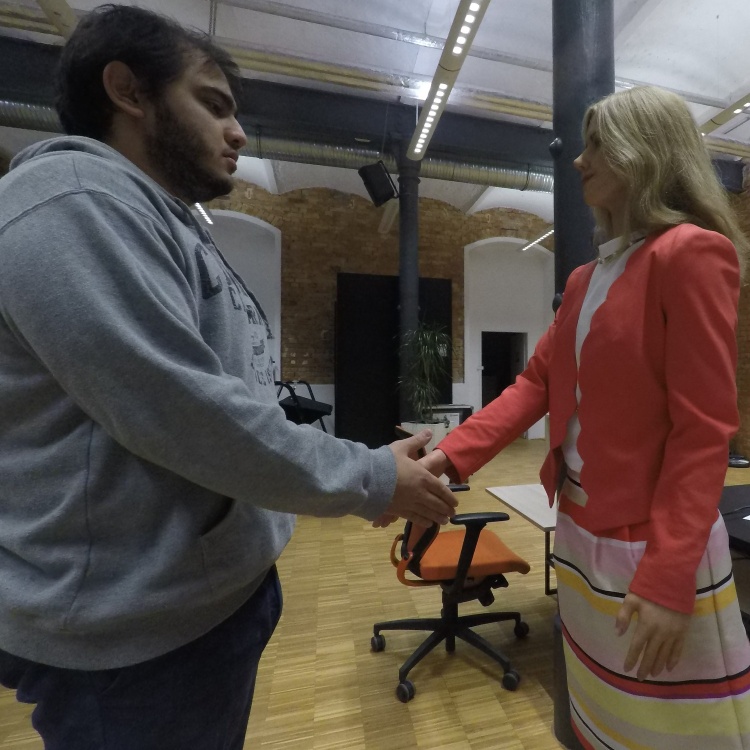}
    % \caption{}
    % \label{fig:sim_pose_0}
    \end{subfigure}%
    \begin{subfigure}[b]{0.24\textwidth}
    \includegraphics[width=\textwidth]{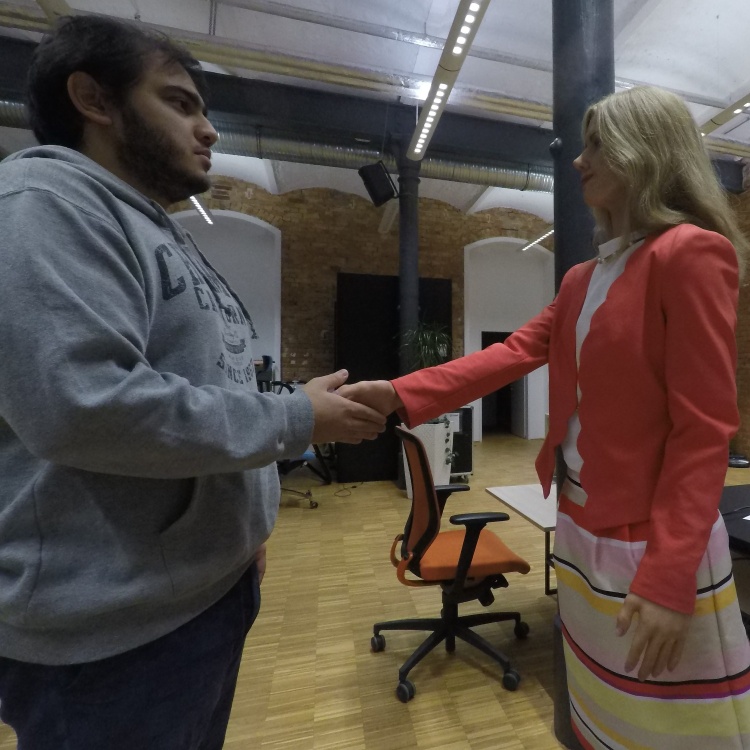}
    % \caption{}
    % \label{fig:sim_pose_25}
    \end{subfigure}%
    \begin{subfigure}[b]{0.24\textwidth}
    \includegraphics[width=\textwidth]{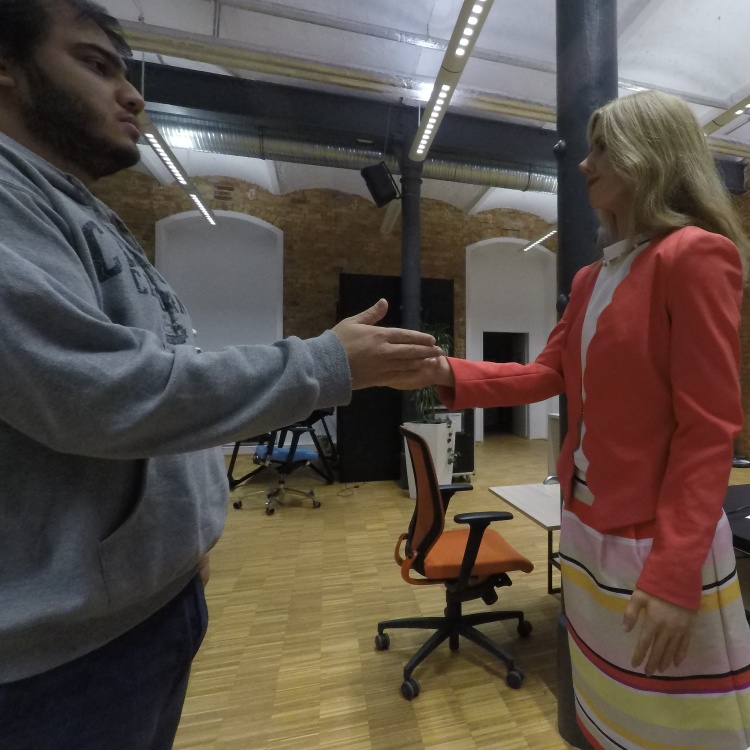}
    % \caption{}
    % \label{fig:sim_pose_50}
    \end{subfigure}%
    \begin{subfigure}[b]{0.24\textwidth}
    \includegraphics[width=\textwidth]{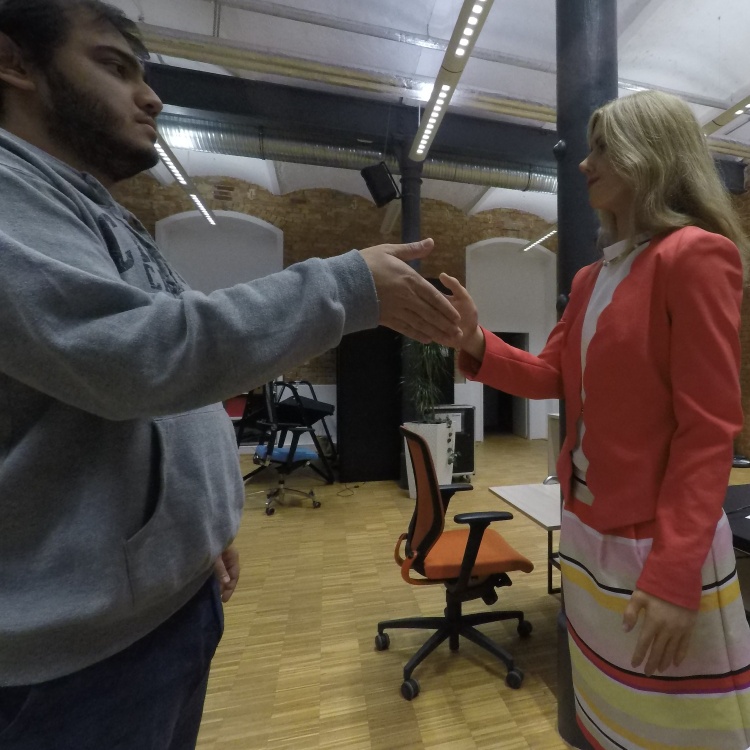}
    % \caption{}
    % \label{fig:sim_pose_75}
    \end{subfigure}%
 \caption{Spatial robustness of the learnt reaching behaviour for different hand locations with different robots.}
 \label{fig:spatial_robustness}
\end{figure*}

Given that we explicitly condition the learnt ProMP on a target 3D location, we attain a good accuracy with the hand reaching behaviours. An example interaction can be seen in Fig. \ref{fig:interaction}, where the timeliness of the interaction can also be seen as the robot starts moving as it senses the human hand moving due to the explicit modelling of the interactive behaviour. Few qualitative samples showing the spatial robustness of our method can be seen in Fig. \ref{fig:spatial_robustness}.  We obtain an average error of $0.062 \pm 0.025$ m of the final hand location and the robot end-effector for Pepper over 10 interaction trajectories. Some errors are attributed to errors in the hand detection and calibration of the setup as well. Due to improper feedback from the pneumatic controller for Elenoide, the average reaching error cannot be properly calculated. 

\section{Conclusion and Future Work}
In our paper, we extend the framework proposed by Vinayavekhin et al. \cite{vinayavekhin2017human} for human-robot handshaking, by learning a robot controller from human interactions. This is especially important when using imitation learning for pneumatically controlled robots that are not backdrivable, like one of the robots used in this paper. The dynamics of the interaction are explicitly modelled using an LSTM network, whose output is used to condition the robot controller to meet the human hand at the predicted location. It does so in a timely and smooth manner. We exploit the fact that humanoid robots have similar degrees of freedom as a human arm and apply the learnt ProMP controller to different humanoid robots directly without any re-training, which is especially tedious if the training requires human interaction partners. 

Currently, we only look at the reaching phase of handshaking. The grasping and shaking phases requires a suitable synergic mechanism that can sense the forces applied and react accordingly. Further research is also required to learn ProMPs for shaking along with reaching, as one is rhythmic, while the other is stroke-based. Sensing the context and reacting accordingly is also important, for example speeding up the motion or strengthening the grasp, could change the way the interaction is perceived. Finally, the true test of how good an interactive behaviour is, would require trials with human partners, which we currently defer to our future work. Additionally, we aim to explore the application of such a method to learning other physically interactive behaviours like high fives, fist bumps etc.

\section{Acknowledgements}
This research was funded by the Interdisciplinary Research Forum (Forum Interdisziplin\"are Forschung - FiF) at TU Darmstadt, the Funding Association for Market-Oriented Management, Marketing, and Human Resource Management (F\"orderverein f\"ur Marktorientierte Unternehmensf\"uhrung, Marketing und Personal management e.V.), the Leap in Time Foundation (Leap in Time Stiftung), and the Center for Responsible Digitization (Zentrum Verantwortungsbewusste Digitalisierung - ZEVEDI). The authors would like to thank Dorothea Koert, Niyati Rawal and Xihao Wang for their useful comments. 

\bibliographystyle{IEEEtran}  % do not change this line!
\bibliography{root}  % put name of your .bib file here

\begin{thebibliography}{10}
\providecommand{\url}[1]{#1}
\csname url@rmstyle\endcsname
\providecommand{\newblock}{\relax}
\providecommand{\bibinfo}[2]{#2}
\providecommand\BIBentrySTDinterwordspacing{\spaceskip=0pt\relax}
\providecommand\BIBentryALTinterwordstretchfactor{4}
\providecommand\BIBentryALTinterwordspacing{\spaceskip=\fontdimen2\font plus
\BIBentryALTinterwordstretchfactor\fontdimen3\font minus
  \fontdimen4\font\relax}
\providecommand\BIBforeignlanguage[2]{{%
\expandafter\ifx\csname l@#1\endcsname\relax
\typeout{** WARNING: IEEEtran.bst: No hyphenation pattern has been}%
\typeout{** loaded for the language `#1'. Using the pattern for}%
\typeout{** the default language instead.}%
\else
\language=\csname l@#1\endcsname
\fi
#2}}

\bibitem{sakamoto2005cooperative}
D.~Sakamoto, T.~Kanda, T.~Ono, M.~Kamashima, M.~Imai, and H.~Ishiguro,
  ``Cooperative embodied communication emerged by interactive humanoid
  robots,'' \emph{International Journal of Human-Computer Studies}, vol.~62,
  no.~2, pp. 247--265, 2005.

\bibitem{hertenstein2006touch}
M.~J. Hertenstein, D.~Keltner, B.~App, B.~A. Bulleit, and A.~R. Jaskolka,
  ``Touch communicates distinct emotions.'' \emph{Emotion}, vol.~6, no.~3, p.
  528, 2006.

\bibitem{yohanan2012role}
S.~Yohanan and K.~E. MacLean, ``The role of affective touch in human-robot
  interaction: Human intent and expectations in touching the haptic creature,''
  \emph{International Journal of Social Robotics}, vol.~4, no.~2, pp. 163--180,
  2012.

\bibitem{han2012robotic}
M.-J. Han, C.-H. Lin, and K.-T. Song, ``Robotic emotional expression generation
  based on mood transition and personality model,'' \emph{IEEE transactions on
  cybernetics}, vol.~43, no.~4, pp. 1290--1303, 2012.

\bibitem{stock2018can}
R.~M. Stock, ``Can service robots hamper customer anger and aggression after a
  service failure?'' Darmstadt Technical University, Department of Business
  Administration~…, Tech. Rep., 2018.

\bibitem{chaplin2000handshaking}
W.~F. Chaplin, J.~B. Phillips, J.~D. Brown, N.~R. Clanton, and J.~L. Stein,
  ``Handshaking, gender, personality, and first impressions.'' \emph{Journal of
  personality and social psychology}, vol.~79, no.~1, p. 110, 2000.

\bibitem{stewart2008exploring}
G.~L. Stewart, S.~L. Dustin, M.~R. Barrick, and T.~C. Darnold, ``Exploring the
  handshake in employment interviews.'' \emph{Journal of Applied Psychology},
  vol.~93, no.~5, p. 1139, 2008.

\bibitem{mori2012uncanny}
M.~Mori, K.~F. MacDorman, and N.~Kageki, ``The uncanny valley [from the
  field],'' \emph{IEEE Robotics \& Automation Magazine}, vol.~19, no.~2, pp.
  98--100, 2012.

\bibitem{vinayavekhin2017human}
P.~Vinayavekhin, M.~Tatsubori, D.~Kimura, Y.~Huang, G.~De~Magistris,
  A.~Munawar, and R.~Tachibana, ``Human-like hand reaching by motion prediction
  using long short-term memory,'' in \emph{International Conference on Social
  Robotics}.\hskip 1em plus 0.5em minus 0.4em\relax Springer, 2017, pp.
  156--166.

\bibitem{campbell2019learning}
J.~Campbell, A.~Hitzmann, S.~Stepputtis, S.~Ikemoto, K.~Hosoda, and H.~B. Amor,
  ``Learning interactive behaviors for musculoskeletal robots using bayesian
  interaction primitives,'' in \emph{2019 IEEE/RSJ International Conference on
  Intelligent Robots and Systems (IROS)}, 2019.

\bibitem{christen2019guided}
S.~Christen, S.~Stevsic, and O.~Hilliges, ``Guided deep reinforcement learning
  of control policies for dexterous human-robot interaction,'' in \emph{2019
  IEEE/RSJ International Conference on Intelligent Robots and Systems (IROS)},
  2019.

\bibitem{prasad2020advances}
V.~Prasad, R.~Stock-Homburg, and J.~Peters, ``Advances in human-robot
  handshaking,'' in \emph{International Conference on Social Robotics}.\hskip
  1em plus 0.5em minus 0.4em\relax Springer, 2020.

\bibitem{prasad2021human}
------, ``Human-robot handshaking: A review,'' \emph{International Journal of
  Social Robotics}, 2021.

\bibitem{jindai2007development}
M.~Jindai and T.~Watanabe, ``Development of a handshake robot system based on a
  handshake approaching motion model,'' in \emph{2007 IEEE/ASME international
  conference on advanced intelligent mechatronics}.\hskip 1em plus 0.5em minus
  0.4em\relax IEEE, 2007, pp. 1--6.

\bibitem{fritsche2015first}
L.~Fritsche, F.~Unverzag, J.~Peters, and R.~Calandra, ``First-person
  tele-operation of a humanoid robot,'' in \emph{2015 IEEE-RAS 15th
  International Conference on Humanoid Robots (Humanoids)}.\hskip 1em plus
  0.5em minus 0.4em\relax IEEE, 2015, pp. 997--1002.

\bibitem{paraschos2013probabilistic}
A.~Paraschos, C.~Daniel, J.~R. Peters, and G.~Neumann, ``Probabilistic movement
  primitives,'' in \emph{Advances in neural information processing systems},
  2013, pp. 2616--2624.

\bibitem{gomez2020adaptation}
S.~Gomez-Gonzalez, G.~Neumann, B.~Sch{\"o}lkopf, and J.~Peters, ``Adaptation
  and robust learning of probabilistic movement primitives,'' \emph{IEEE
  Transactions on Robotics}, vol.~36, no.~2, pp. 366--379, 2020.

\bibitem{hochreiter1997lstm}
S.~Hochreiter and J.~Schmidhuber, ``Lstm can solve hard long time lag
  problems,'' in \emph{Advances in neural information processing systems},
  1997, pp. 473--479.

\bibitem{rasouli2020deep}
A.~Rasouli, ``Deep learning for vision-based prediction: A survey,''
  \emph{arXiv preprint arXiv:2007.00095}, 2020.

\bibitem{sapinski2019emotion}
T.~Sapi{\'n}ski, D.~Kami{\'n}ska, A.~Pelikant, and G.~Anbarjafari, ``Emotion
  recognition from skeletal movements,'' \emph{Entropy}, vol.~21, no.~7, p.
  646, 2019.

\bibitem{metallinou2013tracking}
A.~Metallinou, A.~Katsamanis, and S.~Narayanan, ``Tracking continuous emotional
  trends of participants during affective dyadic interactions using body
  language and speech information,'' \emph{Image and Vision Computing},
  vol.~31, no.~2, pp. 137--152, 2013.

\bibitem{paraschos2018using}
A.~Paraschos, C.~Daniel, J.~Peters, and G.~Neumann, ``Using probabilistic
  movement primitives in robotics,'' \emph{Autonomous Robots}, vol.~42, no.~3,
  pp. 529--551, 2018.

\bibitem{pandey2018mass}
A.~K. Pandey and R.~Gelin, ``A mass-produced sociable humanoid robot: Pepper:
  The first machine of its kind,'' \emph{IEEE Robotics \& Automation Magazine},
  vol.~25, no.~3, pp. 40--48, 2018.

\bibitem{nuitrack}
3DiVi, ``Nuitrack,'' \url{https://nuitrack.com/}, [Online; accessed
  17-Oct-2020].

\bibitem{keselman2017intel}
L.~Keselman, J.~Iselin~Woodfill, A.~Grunnet-Jepsen, and A.~Bhowmik, ``Intel
  realsense stereoscopic depth cameras,'' in \emph{Proceedings of the IEEE
  Conference on Computer Vision and Pattern Recognition Workshops}, 2017, pp.
  1--10.

\bibitem{quigley2009ros}
M.~Quigley, K.~Conley, B.~Gerkey, J.~Faust, T.~Foote, J.~Leibs, R.~Wheeler, and
  A.~Y. Ng, ``Ros: an open-source robot operating system,'' in \emph{ICRA
  workshop on open source software}, vol.~3, no. 3.2.\hskip 1em plus 0.5em
  minus 0.4em\relax Kobe, Japan, 2009, p.~5.

\bibitem{paszke2019pytorch}
A.~Paszke, S.~Gross, F.~Massa, A.~Lerer, J.~Bradbury, G.~Chanan, T.~Killeen,
  Z.~Lin, N.~Gimelshein, L.~Antiga, \emph{et~al.}, ``Pytorch: An imperative
  style, high-performance deep learning library,'' in \emph{Advances in neural
  information processing systems}, 2019, pp. 8026--8037.

\bibitem{kingma2014adam}
D.~P. Kingma and J.~Ba, ``Adam: A method for stochastic optimization,''
  \emph{arXiv preprint arXiv:1412.6980}, 2014.

\bibitem{campbell2017bayesian}
J.~Campbell and H.~B. Amor, ``Bayesian interaction primitives: A slam approach
  to human-robot interaction,'' in \emph{Conference on Robot Learning}, 2017,
  pp. 379--387.

\bibitem{scipy}
P.~{Virtanen}, R.~{Gommers}, T.~E. {Oliphant}, M.~{Haberland}, T.~{Reddy},
  D.~{Cournapeau}, E.~{Burovski}, P.~{Peterson}, W.~{Weckesser}, J.~{Bright},
  S.~J. {van der Walt}, M.~{Brett}, J.~{Wilson}, K.~{Jarrod Millman},
  N.~{Mayorov}, A.~R.~J. {Nelson}, E.~{Jones}, R.~{Kern}, E.~{Larson},
  C.~{Carey}, {\.I}.~{Polat}, Y.~{Feng}, E.~W. {Moore}, J.~{Vand erPlas},
  D.~{Laxalde}, J.~{Perktold}, R.~{Cimrman}, I.~{Henriksen}, E.~A. {Quintero},
  C.~R. {Harris}, A.~M. {Archibald}, A.~H. {Ribeiro}, F.~{Pedregosa}, P.~{van
  Mulbregt}, and S.~.~. {Contributors}, ``{SciPy 1.0--Fundamental Algorithms
  for Scientific Computing in Python},'' \emph{arXiv e-prints}, p.
  arXiv:1907.10121, Jul 2019.

\bibitem{Shahroudy_2016_NTURGBD}
A.~Shahroudy, J.~Liu, T.-T. Ng, and G.~Wang, ``Ntu rgb+d: A large scale dataset
  for 3d human activity analysis,'' in \emph{IEEE Conference on Computer Vision
  and Pattern Recognition}, June 2016.

\bibitem{zhang2012microsoft}
Z.~Zhang, ``Microsoft kinect sensor and its effect,'' \emph{IEEE multimedia},
  vol.~19, no.~2, pp. 4--10, 2012.

\end{thebibliography}

\end{document}